\documentclass[sigconf]{acmart}

\usepackage{multirow}
\usepackage{balance}

\AtBeginDocument{%
  \providecommand\BibTeX{{%
    \normalfont B\kern-0.5em{\scshape i\kern-0.25em b}\kern-0.8em\TeX}}}

\copyrightyear{2021}
\acmYear{2021}
\setcopyright{acmcopyright}\acmConference[MM '21]{Proceedings of the 29th ACM International Conference on Multimedia}{October 20--24, 2021}{Virtual Event, China}
\acmBooktitle{Proceedings of the 29th ACM International Conference on Multimedia (MM '21), October 20--24, 2021, Virtual Event, China}
\acmPrice{15.00}
\acmDOI{10.1145/3474085.3475479}
\acmISBN{978-1-4503-8651-7/21/10}

\begin{document}
\fancyhead{}
\title{Long-tailed Distribution Adaptation}


\author{Zhiliang Peng,\quad Wei Huang,\quad Zonghao Guo,\quad Xiaosong Zhang,\quad Jianbin Jiao,\quad Qixiang Ye$^{*}$}

 \makeatletter
 \def\authornotetext#1{
 \if@ACM@anonymous\else
     \g@addto@macro\@authornotes{
     \stepcounter{footnote}\footnotetext{#1}}
 \fi}
 \makeatother
 \authornotetext{Corresponding author.}


\affiliation{
 \institution{University of Chinese Academy of Sciences}
 \city{Beijing}
 \country{China}
}
\email{{pengzhiliang19, huangwei19, guozonghao19, zhangxiaosong18}@mails.ucas.ac.cn,
{jiaojb, qxye}@ucas.ac.cn}

\def\authors{Zhiliang Peng, Wei Huang, Zonghao Guo, Xiaosong Zhang, Jianbin Jiao, Qixiang Ye}

\renewcommand{\shortauthors}{Peng et al.}


\begin{abstract}
  Recognizing images with long-tailed distributions remains a challenging problem while there lacks an interpretable mechanism to solve this problem. In this study, we formulate Long-tailed recognition as Domain Adaption (LDA), by modeling the long-tailed distribution as an unbalanced domain and the general distribution as a balanced domain. Within the balanced domain, we propose to slack the generalization error bound, which is defined upon the empirical risks of unbalanced and balanced domains and the divergence between them. We propose to jointly optimize empirical risks of the unbalanced and balanced domains and approximate their domain divergence by intra-class and inter-class distances, with the aim to adapt models trained on the long-tailed distribution to general distributions in an interpretable way. Experiments on benchmark datasets for image recognition, object detection, and instance segmentation validate that our LDA approach, beyond its interpretability, achieves state-of-the-art performance. Code
is available at \href{https://github.com/pengzhiliang/LDA}{\color{magenta}github.com/pengzhiliang/LDA}.
\end{abstract}

\ccsdesc[500]{Computing methodologies~Cost-sensitive learning}
\ccsdesc[300]{Computing methodologies~Image representations}
\ccsdesc[100]{Computing methodologies~Object detection}

\keywords{long-tail distribution, domain adaptation, classification, object detection, instance segmentation}


\maketitle

\section{Introduction}
The success of deep learning depends on large-scale high-quality datasets, \textit{e.g.}, ImageNet ILSVRC 2012 \cite{russakovsky2015imagenet}, MS COCO \cite{lin2014microsoft} and Places Database \cite{zhou2017places}, where the data uniformly distributes with respect to categories. The real-world data, however, typically follows long-tailed distributions~\cite{cui2018large,gupta2019lvis}, \textit{i.e.}, a few classes (\textit{a.k.a.} head classes) occupy most of the data while most classes (\textit{a.k.a.} tail classes) have few samples. 

To adapt the models trained upon long-tailed distributions to uniform distributions, re-sampling and re-weighting methods~\cite{buda2018systematic,cui2019class} have been explored to re-balance the data distributions. 
Recently, feature-classifier decoupling learning strategies~\cite{kang2020decoupling, zhou2020bbn, jamal2020rethinking, ren2020balanced} are proposed to learn representation and classifiers, step by step. Despite of the effectiveness, these approaches require complicated training strategies and/or work in empirical fashions, which hinder the interpretability and further progress of long-tailed recognition problems. 

In this study, analogous to~\cite{jamal2020rethinking}, we treat the unbalanced training set as an unbalanced domain and the balanced test set as a balanced domain, solving the Long-tailed recognition problem in a Domain Adaptation (LDA) fashion.
Based on the upper bound of generalization error of domain adaptation~\cite{ben2007analysis, ben2010theory}, LDA firstly defines the relaxed upper bound of generalization error for the balanced domain. The training objective of LDA is to minimize the defined relaxed upper bound, which is composed of empirical risks of the unbalanced domain and balanced domain and the divergence between these two domains.
By the way, the proposed relaxed upper bound has a great potential in explaining popular decoupling learning strategies~\cite{kang2020decoupling, zhou2020bbn, jamal2020rethinking, ren2020balanced}.
\begin{figure}[!t]
\begin{center}
 \includegraphics[width=1.0\linewidth]{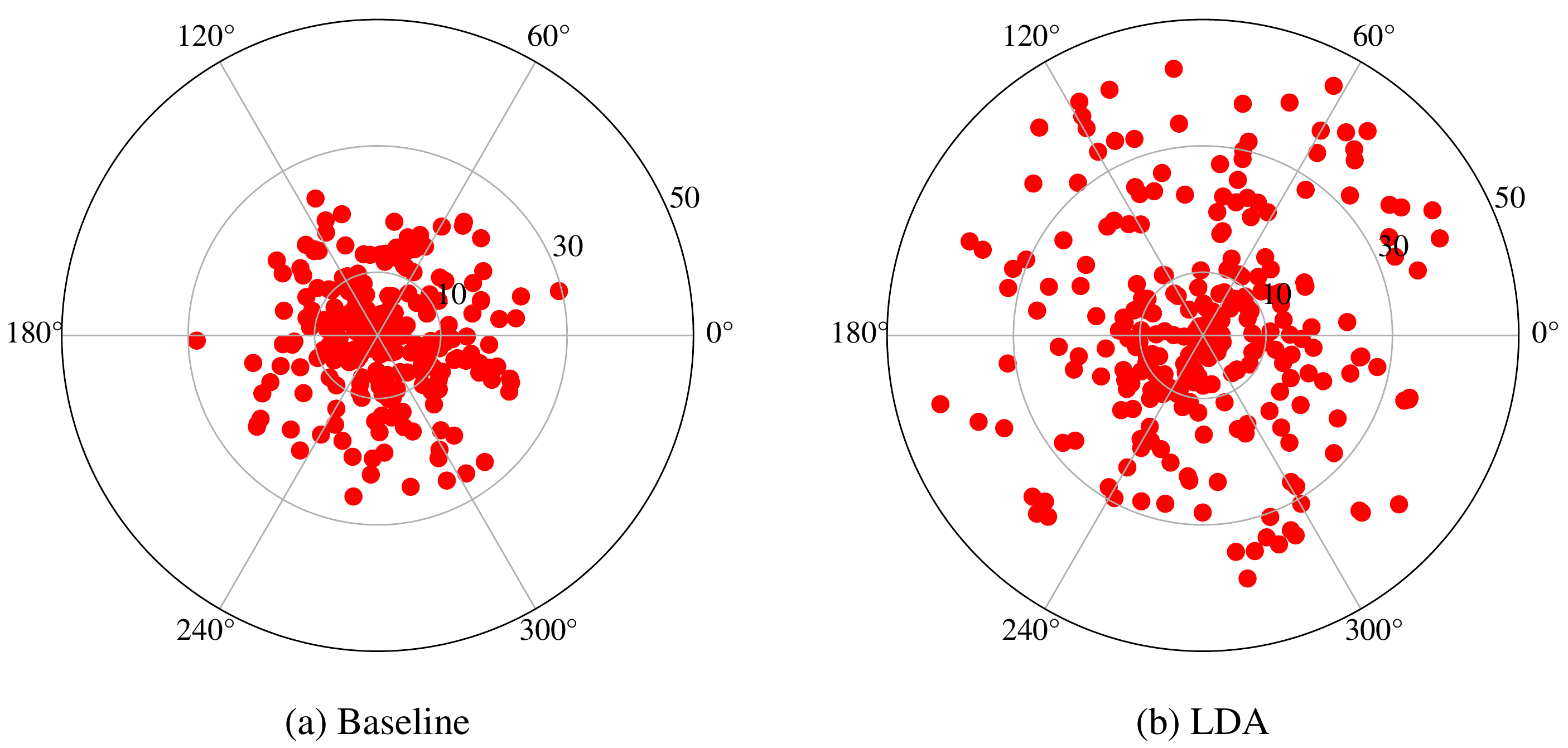}
\end{center}
 \caption{t-SNE visualization of classifier weight vectors. (a) Classifier weight vectors trained with the cross entropy loss, and (b) with the proposed LDA approach. Larger distances among vectors imply higher discriminative capacity.}
\label{fig:tsne_weight} 
\end{figure}

During training, LDA minimizes the empirical risk of the unbalanced and balanced domains in an end-to-end manner. Considering that balanced and unbalanced domains follow the same distributions but have different sampling ratios, empirical risk of the balanced domain is estimated by using the sample re-weighting method~\cite{zhang2013domain}. However, it is challenging for a classifier $h$ to simultaneously minimize two homogeneous but different risks (balanced and unbalanced empirical risks). Therefore, an auxiliary classifier $h'$, is introduced to collaborate with $h$ and minimize the unbalanced and the balanced empirical risks, respectively.

Meanwhile, a dynamic balance factor defined on the difference between the two empirical risks is used to regularize the two risks, which prevents the model from favoring either risk. Compared with recent works (including decoupling learning strategy~\cite{kang2020decoupling} and BBN~\cite{zhou2020bbn}) which focus on tail classes when the network tends to converge, our LDA approach consistently considers tail and head classes during the entire training procedure, which can prevent the network from falling into a local minimum.

\begin{figure*}[t]
\begin{center}
 \includegraphics[width=1\linewidth]{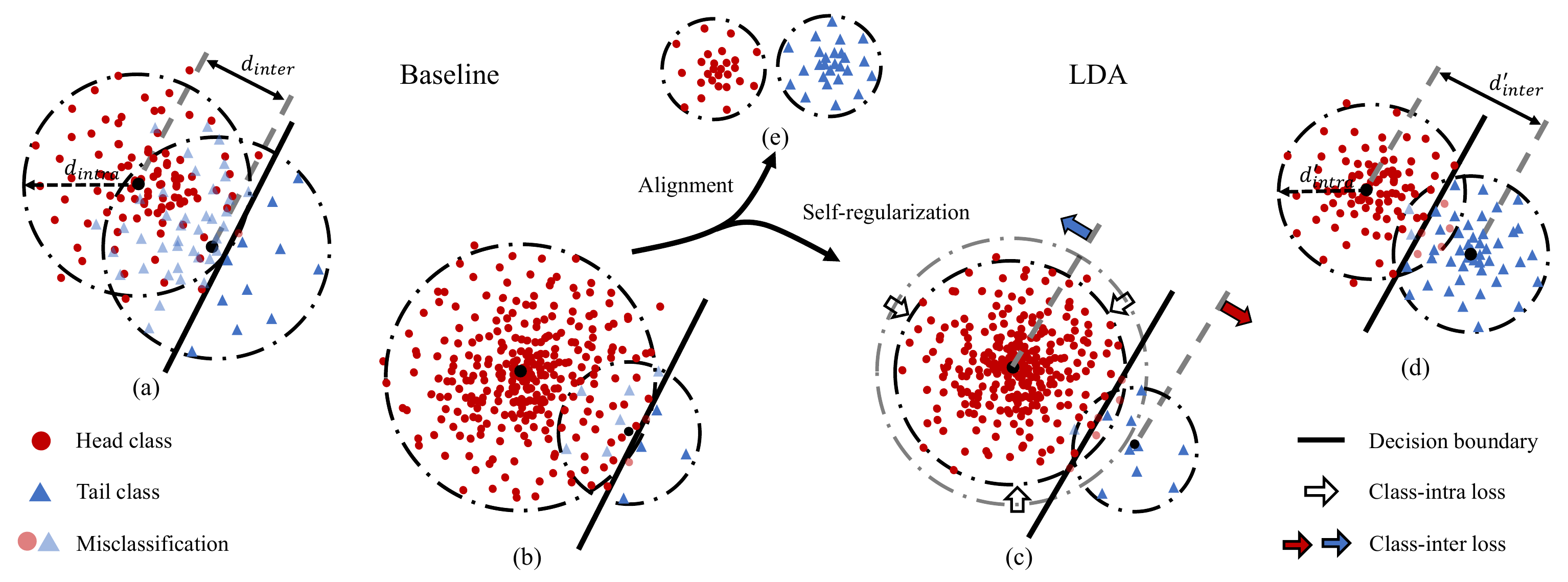}
\end{center}
 \caption{
Overview of the proposed LDA approach. (a) and (b) are the feature spaces of balanced and unbalanced domains generated by the baseline model, which is trained with the cross entropy loss. The model biases towards head classes as their loss dominates the unbalanced empirical risk, $e.g.,$ the classifier and generated feature distribution (c) and (d) respectively denote feature spaces of the unbalanced and balanced domains generated by our LDA approach. The model has significantly smaller bias towards head classes. (e) is the feature space of the ideal balanced domain as the alignment goal for unbalanced domain. (Best viewed in color)
}
\label{fig:scatter}
\end{figure*}

LDA involves a plausible strategy to reduce the divergence between balanced and unbalanced domains. The primary difference between long-tail recognition with general domain adaptation is that the divergence between balanced and unbalanced domains is not observable. We therefore propose to minimize the divergence by constraining feature distribution of the unbalanced domain, Figure~\ref{fig:scatter}(c). Specially, we construct an ideal balanced domain, where features of each class have minimal overlap, as the learning objective, Figure~\ref{fig:scatter}(e). Furthermore, a self-regularization strategy that increases inter-class distance and decreases intra-class distance is proposed to reduce the divergence between the unbalanced domain and the ideal balanced domain. Consequently, the divergence, as defined in ~\cite{kang2019contrastive}, has been calculated and shown the effectiveness of the self-regularization strategy.

We evaluate LDA on long-tailed tasks including image classification, object detection and instance segmentation, and validate its superiority over competing approaches. The last but not the least, LDA has a negligible computational cost overhead compared with the baseline method, which is trained with the cross entropy loss.

\section{Related Work}
\subsection{Long-tailed Recognition}
\textbf{Re-balancing Method.} This line of methods can be categorized to re-sampling and re-weighting ones. The re-sampling methods are further categorized into two: 1) Over-sampling by simply repeating data for minority classes~\cite{shen2016relay, buda2018systematic, byrd2019effect, gupta2019lvis} and 2) under-sampling by abandoning data for dominant classes~\cite{japkowicz2002class,he2009learning,buda2018systematic}. 
The re-weighting methods usually assign larger weights to samples from tail classes in the loss function~\cite{huang2016learning, cui2019class, byrd2019effect} or drop some scores of tail classes in the Softmax function to balance the positive and negative gradients ~\cite{tan2020equalization}. 

Most re-balancing strategies try to restore the balanced distribution from the unbalanced set, so as to directly minimize the balanced empirical risk. Nevertheless they unfortunately ignore the unbalanced empirical risk and distort the feature distribution to some extent.

\noindent\textbf{Decoupling Method.} Considering that the balanced and unbalanced data share a feature space but different distributions, various decoupling strategies~\cite{ouyang2016factors, cui2018large, cao2019learning, kang2020decoupling, jamal2020rethinking, ren2020balanced} separated the training process into two stages. In the first stage, they train the network upon the unbalanced data to construct the feature space. In the second stage, they utilize re-balancing strategies to ﬁne-tune the classifiers with small learning rates to elaborate the classification boundaries. The representative BBN method~\cite{zhou2020bbn} has a two-branch architecture, which equips with two samplers and shifts attention from head classes to tail classes during the training process. 

Despite of the effectiveness, these methods either significantly increase the parameters or require a complicated training strategy. The proposed LDA overcomes these disadvantages by training a simpler yet more effective paradigm, as well as considering the divergence between two domains for distribution alignment.

\subsection{Domain Adaptation}

Domain adaptation methods aim to mitigate the divergence between the distribution of training data and test data so that the learned models can be well generalized to target domains~\cite{shimodaira2000improving, gong2012geodesic, zhang2019a}. 
There are some approaches that handle the imbalance problem during domain adaptation ~\cite{hongliang2017mind, zou2018unsupervised}. The test data in the domain adaptation setting can be used to explicitly calculate divergence, which is however invisible for long-tailed recognition during the training procedure. One recent method~\cite{jamal2020rethinking} estimated the conditional probability within balanced empirical risk by the meta-learning method and adopt the decoupling training strategy. However, the divergence between the unbalanced training set and the balanced test set remains overlooked and requires to be elaborated.

\section{Methodology}

\subsection{Modeling}
We formulate long-tailed recognition as a domain adaptation task, where the source domain corresponds to a unbalanced domain (the unbalanced training data distribution) and the target domain to a balanced domain (the balanced test data distribution). Denoting the two domains as $\mathcal{D}_{u}$ and $\mathcal{D}_{b}$, their distributions in feature space are defined as $\tilde{\mathcal{D}}_{u}$ and $\tilde{\mathcal{D}}_{b}$, respectively.
Notably, we not only consider the long-tailed recognition problem as a domain adaptation task,  analogous to~\cite{jamal2020rethinking}, but also propose to define and slack the  generalization error bound.

In domain adaptation, generalization error upper bound of the target domain has been extensively studied~\cite{ben2007analysis,ben2010theory}. Let $\mathcal{R}$ be a representation function from the input sample space $\mathcal{X}$ to the feature space $\mathcal{Z}$ and $\mathcal{H}$ a hypothesis space with \textit{VC-dimension} $d$. If a random labeled sample set of size $m$ is generated by applying $\mathcal{R}$ to a $\mathcal{D}_{u}$-i.i.d. labeled samples, then with probability $1- \delta$ at least for every $h \in \mathcal{H}$ the expect risk of the balanced domain $\epsilon_{b}(h)$ satisfies the following inequality:
\begin{equation}
    \begin{split}
        \epsilon_{b}(h) \leq \hat{\epsilon}_{u}(h) &+
        \sqrt{\frac{4}{m}\left(d \log \frac{2 e m}{d} +\log \frac{4}{\delta}\right)} \\
        &+ d_{\mathcal{H}}(\tilde{\mathcal{D}}_{u}, \tilde{\mathcal{D}}_{b})
        +\lambda,
    \end{split}
\label{eq1}
\end{equation}
\noindent where $\hat{\epsilon}_{u}(h)$ denotes empirical risk of the unbalanced domain and $\lambda$ is a constant. 
$d_{\mathcal{H}}(\tilde{\mathcal{D}}_{u}, \tilde{\mathcal{D}}_{b})$ specifies the divergence between $\tilde{\mathcal{D}}_{u}$ and $\tilde{\mathcal{D}}_{b}$.

Based on the $VC$ theory~\cite{vapnik1999overview}, expect risk $\epsilon_{b}(h)$ is bound to its empirical estimation $\hat{\epsilon}_{b}(h)$. Namely, if $b$ is an $m'$-size i.i.d. sample, then with probability exceeding $1 - \delta$, $\epsilon_{b}(h)$ satisfies the following inequality:
\begin{equation}
    \epsilon_{b}(h) \leq \hat{\epsilon}_{b}(h)+
    \sqrt{\frac{4}{m'}\left(d \log \frac{2 e m'}{d}+
    \log \frac{4}{\delta}\right)}.
\label{eq2}
\end{equation}
Notably, images in an unbalanced training set and balanced validation/test set are sampled from the same distribution but with different sample ratios. It remains satisfying the hypothesis that $\mathcal{D}_{b}$ is contained by $\mathcal{D}_{u}$ ~\cite{zhang2013domain}. Therefore, $\hat{\epsilon}_{b}(h)$ can be estimated upon $\hat{\epsilon}_{u}(h)$:
\begin{equation}
    \begin{split}
        \hat{\epsilon}_{b}(h)  &= \mathbb{E}_{p_{b}(x, y)}\hat{\epsilon}_{u}(h(x), y) \\
        &= \mathbb{E}_{p_{u}(x, y)}\frac{p_{b}(x, y)}{p_{u}(x, y)}\hat{\epsilon}_{u}(h(x), y) \\
    \end{split}
\label{eq3}
\end{equation}
\noindent where $\mathbb{E}$ is expectation function and $p_{u}(x, y)$ and $p_{b}(x, y)$ are joint probability in unbalanced and balanced domain, respectively.

So far, $\epsilon_{b}(h)$ is confined to two upper bounds defined in Equations \ref{eq1} and \ref{eq2}. However, it is nontrivial to determine its strict minimal upper bound in theory and we propose to approximate an upper bound by
\begin{equation}
    \epsilon_{b}(h) \lesssim \hat{\epsilon}_{u}(h)+ \hat{\epsilon}_{b}(h) + d_{\mathcal{H}}\left(\tilde{\mathcal{D}}_{u}, \tilde{\mathcal{D}}_{b}\right)+\lambda^{*},
\label{slackupper}
\end{equation}
where $\lambda^{*}$ is the constant term.

As shown in Equation \ref{slackupper}, expect risk of the balanced domain $\epsilon_{b}(h)$ is connected with empirical risks of the unbalanced domain $\hat{\epsilon}_{u}(h)$ and the balanced domain $\hat{\epsilon}_{b}(h)$ and the divergence $d_{\mathcal{H}}(\tilde{\mathcal{D}}_{u}, \tilde{\mathcal{D}}_{b})$ between two domains. In what follows, we propose to optimize these items one by one while minimizing $\epsilon_{b}(h)$.


\subsection{Empirical Risk Minimization}\label{minimizingrisks}
With a single classifier $h$ (termed balanced classifier), it is difficult to minimize two homogeneous risk functions defined in Equation \ref{slackupper}. We therefore introduce an auxiliary classifier $h'$ (termed unbalanced classifier) to minimize $\hat{\epsilon}_{u}$. In this way, we relax the contraint for the balanced classifier $h$, by solely minimizing $\hat{\epsilon}_{b}$, without reducing $\hat{\epsilon}_{u}$ explicitly. 

Notably, the representation function $\mathcal{R}$ in \cite{ben2007analysis, ben2010theory} is assumed be fixed, but it's learnable in the CNN framework with gradient comes from the classifier.
If we regard the whole network as a complex classifier $\theta$ (including $\mathcal{R}$, $h$ and $h'$), $\theta$ should also subject to the slack upper bound defined by Equation~\ref{slackupper}.
In this case, $\mathcal{R}$ can be optimized regardless of which classifier the empirical risk comes from.
Therefore, introducing $h'$ is plausible, as it relaxes the constraint on $h$ but does not affect the optimization objective of $\theta$.

Based on the above analysis, we conclude the risk/loss function for $h$ and $h'$. For $h'$, it requires to minimize empirical risk for the unbalanced domain, \textit{i.e.}, $\hat{\epsilon}_{u}(\mathcal{R}, h') = \frac{1}{N} \sum_{i=1}^{N}\mathcal{L}(f(x_{i}, \mathcal{R}, h'), y_{i})$, where $N$ denotes the samples number in a batch and $\mathcal{L}$ is the vanilla cross entropy loss.
For $h$, we would like to have a closer look at $\hat{\epsilon}_{b}(\mathcal{R}, h)$, 
from the view of target shift case in domain adaptation~\cite{zhang2013domain}.
In this case, the conditional probability is shared between balanced and unbalanced domains, \textit{i.e.}, $p_{u}(x|y) = p_{b}(x|y)$, which is different from~\cite{jamal2020rethinking}, while the marginal probability is different, \textit{i.e.}, $p_{u}(y) \ne p_{b}(y)$. Therefore, the estimated $\hat{\epsilon}_{b}$ is re-defined as: 
\begin{equation}
    \begin{split}
        \hat{\epsilon}_{b}(\mathcal{R}, h)  &= \mathbb{E}_{p_{u}(x, y)}\frac{p_{b}(x, y)}{p_{u}(x, y)}\mathcal{L}(f(x, \mathcal{R}, h), y) \\
        &= \mathbb{E}_{p_{u}(x, y)}\frac{p_{b}(y)p_{b}(x|y)}{p_{u}(y) p_{u}(x|y)}\mathcal{L}(f(x, \mathcal{R}, h), y) \\
        &= \mathbb{E}_{p_{u}(x, y)}\frac{p_{b}(y)}{p_{u}(y)}\mathcal{L}(f(x, \mathcal{R}, h), y) \\
        &= \frac{1}{N} \sum_{i=1}^{N} w_{y_{i}}\mathcal{L}(f(x_{i}, \mathcal{R}, h), y_{i}),
    \end{split}
\label{eq:balancerisk}
\end{equation}
\noindent where $w_{y_{i}}=p_{b}(y_i)/p_{u}(y_i)$. Elaborate designs for $w_{y_{i}}$ can be found in the literature~\cite{lin2017focal, cui2019class} but we implement it as $w_{y_{i}}=1/p_{u}(y_i)$ for simplicity.

\subsection{Empirical Risk Balance}\label{balancingrisk}

Two empirical risks jointly determine the optimization of the network while $\mathcal{R}$ receives supervised information to minimize the risks. If either of them dominates the training process, the optimization tends to get stuck to a local minimum. Therefore, how to balance two empirical risks remains critical.

We propose a self-adaptive adjustment method, based on the real-time performance difference between $h$ and $h'$ during training to balance the two empirical risks. For image $x_i$, let $\hat{y}_{h'i}$ and $\hat{y}_{hi}$ denote the predict labels of classifier $h'$ and $h$ respectively, \textit{i.e.},$\hat{y}_{h'i} = \arg \max f(x_i; \mathcal{R}, h')$, $\hat{y}_{hi} = \arg \max f(x_i; \mathcal{R}, h)$. Let parameter $\alpha$ regularize the importance of $\hat{\epsilon}_{u}(h')$ and $\hat{\epsilon}_{b}(h)$. The empirical risk is a trade-off between $\hat{\epsilon}_{u}(h')$ and $\hat{\epsilon}_{b}(h)$: $\hat{\epsilon}_{u}(h') + \alpha \hat{\epsilon}_{b}(h)$. Let $\Delta acc$ denote the performance differences. $\alpha$ is defined as:
\begin{equation}
    \begin{split}
        \alpha &= \Delta {acc}^{\gamma} \\
        &= (\frac{1}{L}\frac{1}{N}\sum_{l=1}^{L}\sum_{l=1}^{N}(\mathbb{I}[\hat{y}_{hi} \ne y_i] - \mathbb{I}[\hat{y}_{h'i} \ne y_i]))^{\gamma},
    \end{split}
\label{eq:alpha}
\end{equation}
\noindent where $L$ is the sample number in a few previous mini-batches. It avoids $\Delta acc$ changing dramatically. $\mathbb{I}[\hat{y} \ne y_i] \in \{0, 1\}$ is an indicator function evaluating to 1 if $\hat{y} \ne y_i$ else 0. $\gamma$ is an hyper-parameter which controls the degree of self-adaptive adjustment.

Different from the decoupling methods~\cite{cao2019learning, kang2020decoupling}, which optimize the head classes at early training epochs and optimize tail classes after networks are adequately trained, our approach treats all classes equally during the whole training process. 
The self-adaptive adjustment strategy dynamically balances two empirical risks, Figure \ref{fig:ablation-alpha}(b), as: a) When $\Delta acc$ increases $\alpha$ increases, because $\alpha$ is proportional to $\Delta acc$; b) When $\alpha$ increases the network pays more attention to $\hat{\epsilon}_{b}(h)$, driving $\Delta acc$ to decrease; c) when $\Delta acc$ decreases $\alpha$ decreases for the proportional relation; d) when $\alpha$ decreases, the network reduces its attention to $\hat{\epsilon}_{b}(h)$, so that $\Delta acc$ increases. These processes iterate until the convergence of the training procedure.

\begin{table}[!t]
\caption{Classification accuracy on ImageNet-LT. $\dagger$ means copied from \cite{tang2020long}.}
\begin{center}
\begin{tabular}{c|ccc|c}
\toprule
Methods & Few & Medium & Many & All  \\ \midrule
$\text{Focal}^{\dagger}$\cite{lin2017focal} & 8.2 & 37.1 & 64.3 & 43.7 \\
$\text{OLTR}^{\dagger}$\cite{liu2019large} & 20.8 & 40.8 & 51.0& 41.9 \\ \midrule
$\text{NCM}$\cite{kang2020decoupling} & 28.1 & 45.3 & 56.6 & 47.3 \\ 
$\text{cRT}$\cite{kang2020decoupling} & 27.4 & 46.2 & 61.8 & 49.6 \\ 
$\tau\text{-norm}$\cite{kang2020decoupling} & 30.7 & 46.9 & 59.1 & 49.4 \\ 
$\text{LWS}$\cite{kang2020decoupling} & 30.3 & 47.2 & 60.2 & 49.9 \\ 
\midrule
$\text{De-confound}$\cite{tang2020long} & 14.7 & 42.7 & 67.9 & 48.6 \\ 
$\text{De-confound-TDE}$\cite{tang2020long} & \textbf{31.6} & 48.8 & 62.7 & 51.8 \\ \midrule
Baseline & 11.1& 41.6& \textbf{68.8}& 47.9\\ 
(Ours) LDA & 31.5 & \textbf{50.9} & 64.5 & \textbf{53.4} \\ \bottomrule
\end{tabular}
\end{center}
\label{tab:imagenet}
\end{table}

\subsection{Divergence Minimization}\label{minimizingdivergence}
With respect to divergence, the differences between general domain adaptation and long-tailed recognition are twofold. On the one hand, $\tilde{\mathcal{D}}_{b}$ is invisible in long-tailed recognition because images in validation/test set are inaccessible during training. On the other hand, we can obtain an approximately balanced classifier when minimizing the balanced empirical risk, as shown in Figure~\ref{fig:scatter}(d). Different from general domain adaptation, where transfers the features of the target domain to the source domain to guarantee the classifier trained within the source domain yields accurate results in the target domain, the proposed LDA requires to transfer the features of the unbalanced domain to the balanced domain on the contrary.

Considering the difficulty to estimate the hidden space $\tilde{\mathcal{D}}_{b}$, we propose to approach it in an indirect fashion. Assuming there exists an ideal balanced domain space $\tilde{\mathcal{D^{*}}}_{b}$, as shown in Figure~\ref{fig:scatter}(e), which is the alignment objective of the unbalanced domain. Unfortunately, $\tilde{\mathcal{D^{*}}}_{b}$ can't be defined explicitly. We thereby propose to take advantage of the nature of $\tilde{\mathcal{D^{*}}}_{b}$, where the feature space of each class has minimal intersection.
In specific, we propose a self-regularization strategy to approximate the divergence $d_{\mathcal{H}}\left(\tilde{\mathcal{D}}_{u}, \tilde{\mathcal{D}}_{b}\right)$ by minimizing the intersection space among classes. This is implemented by increasing the inter-class distance and decreasing the intra-class distance, as shown in Figure~\ref{fig:scatter}(c).

Based on above analysis, we propose to separate features of different classes and pull together features of same classes to the largest extent. In specific, we add a projection head $p$, \textit{i.e.} a fully connect layer, to project the representation before the classifier to a latent space of higher dimension. To increase the distance between class in the latent space, the feature centers of different classes are required to stay away from each other. 

Let $\mathcal{F}_i$ be the feature of $x_{i}$ in the latent space, $\mu_{c}$ the feature center of class $c$, $i.e.$, $\mu_{c} = \frac{1}{|c|}\sum_{i \in \{ y = c\}} \mathcal{F}_i$, where $|c|$ denotes the feature number for class $c$. The intra-class distance of class $c$ is defined as $d_{intra}^{c} = \frac{1}{|c|}\sum_{i \in \{ y=c\}} 1 - cos(\mathcal{F}_i, \mu_c)$, where $cos$ is the $cosine$ funciton. Similarly, the inter-class distance between class $i$ and $j$ is defined as $d_{inter}^{ij} = 1 - cos(\mu_i, \mu_j)$.

By defining $C_b$ as the number of class appears in a mini-batch, the intra-distance loss $\mathcal{L}_{intra}$, which aims to reduce $d_{intra}^{c}$, is defined as:
\begin{equation}
    \mathcal{L}_{intra} = \frac{1}{C_b}\sum_{c=1}^{C_b} \frac{1}{|c|} \sum_{i \in \{ y=c\}} 1 - cos(\mathcal{F}_i, \mu_c).
\label{intra-loss}
\end{equation}
Minimizing the inter-distance loss $\mathcal{L}_{inter}$ aims to maximize $d_{inter}^{ij}$. It is a variant of hinge loss with margin $\Delta_m$, which is 1 in our experiments unless otherwise noted. $\mathcal{L}_{inter}$ is defined as:
\begin{equation}
    \mathcal{L}_{inter} = \sum_{i=1}^{C_b}\sum_{j=1, j \ne i}^{C_b}\omega_{ij}\max (0, \Delta_m - (1 - cos(\mu_i, \mu_j))),
\label{inter-loss}
\end{equation}
where $\omega_{ij}=w_{i}+w_{j}$, $w_{i}$ and $w_{j}$, which are defined in Equation \ref{eq:balancerisk}, are respectively the weight for class $i$ and $j$. 

\begin{figure}[t]
\begin{center}
\centerline{\includegraphics[width=1\columnwidth]{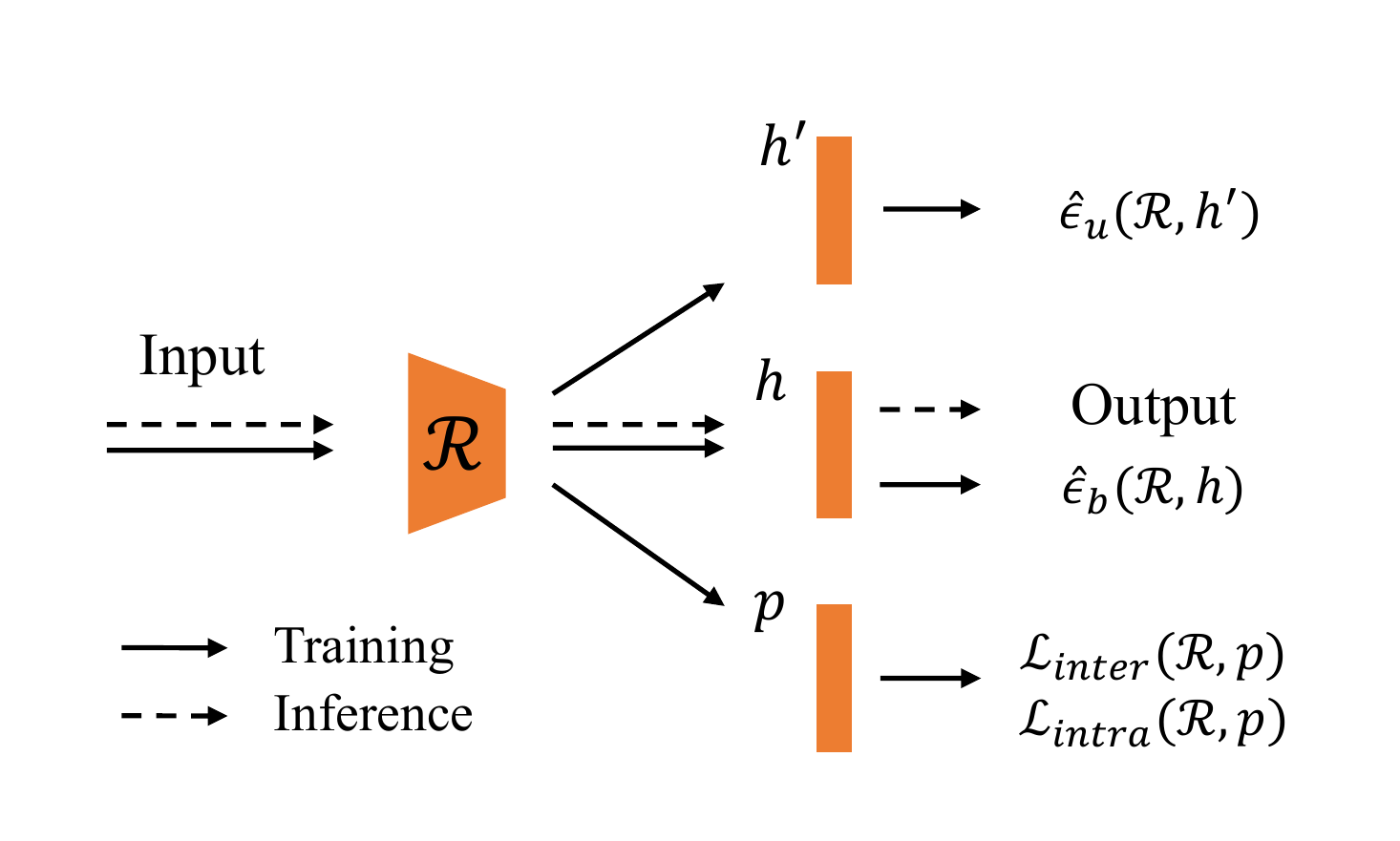}}
\caption{Network architecture. $h'$, $h$, and $p$ are jointly optimized classifiers (fully-connect layers). During inference, $h$ predicts classification results while $h'$ and $p$ do not predict.}
\label{fig-flowchart}  
\end{center}
\end{figure}

\subsection{Long-tail Recognition}

The implementation of long-tailed recognition is illustrated in Figure \ref{fig-flowchart}.
In the training phase, we minimize $\hat{\epsilon}_{u}(\mathcal{R}, h')$ and $\hat{\epsilon}_{b}(\mathcal{R}, h)$ to optimize the unbalanced classifier $h'$ and balanced classifier $h$, respectively. We also minimize $\mathcal{L}_{intra}(\mathcal{R}, p)$ and $\mathcal{L}_{inter}(\mathcal{R}, p)$ to optimize projection head $p$. The total loss function is concluded as:
\begin{equation}
\begin{split}
    \mathcal{L}_{LDA} &= \hat{\epsilon}_{u}(\mathcal{R}, h') + \alpha \hat{\epsilon}_{b}(\mathcal{R}, h) \\
    &+ \beta(\mathcal{L}_{intra}(\mathcal{R}, p) + \mathcal{L}_{inter}(\mathcal{R}, p)),
\end{split}
\label{eq:totalloss}
\end{equation}
\noindent where $\alpha$ is defined in Equation \ref{eq:alpha} and $\beta$ is the weight of $\mathcal{L}_{intra}$ and $\mathcal{L}_{inter}$ penalty term, which is set to 1 in experiments unless otherwise noted.

In the test phase, we utilize the output from balanced classifier to evaluate performance because the test set is a balanced domain. As there is not additional parameters introduced, the proposed LDA method has negligible computational cost in the inference phase compared with the baseline method, which is trained with the cross entropy loss.


\begin{table}[!t]
\caption{Classification accuracy on Long-tailed CIFAR-100.\textbf{*} means using pre-trained models~\cite{tang2020long}.}
\begin{center}
\begin{tabular}{cccc}
\toprule
\multirow{2}{*}{Methods} & \multicolumn{3}{c}{Imbalance Ratios} \\ 
 & 10 & 50 & 100 \\ 
\toprule
Focal\cite{lin2017focal} & 55.8& 44.3& 38.4 \\ 
CB-Focal\cite{cui2019class} & 58.0& 45.3& 39.6\\ 
CE-DRW\cite{cao2019learning} & 58.1& 45.3& 41.5 \\ 
CE-DRS\cite{cao2019learning} & 58.1& 45.5& 41.6 \\ 
LDAM-DRW\cite{cao2019learning} & 58.7& 46.6& 42.0 \\ 
BBN\cite{zhou2020bbn} & 59.2& 47.0& 42.6\\ \midrule 
De-confound\textbf{*}\cite{tang2020long} & 59.5& 48.9& 43.9 \\
De-confound-TDE\textbf{*}\cite{tang2020long} & 59.8& 51.2& 47.3 \\ \midrule
Baseline & 55.7 & 43.9 & 38.4 \\
(Ours) LDA & 59.7 &48.1 & 43.8 \\
(Ours) LDA\textbf{*} &\textbf{61.9}  &\textbf{54.6} & \textbf{50.6} \\
\bottomrule
\end{tabular}
\end{center}
\label{tab:cifar100}
\end{table}

\section{Experiment}
\subsection{Datasets and Setup}
We evaluate the proposed approach on four datasets including the CIFAR-100-LT \cite{cui2019class}, ImageNet-LT \cite{liu2019large},  Places-LT \cite{liu2019large}, and the LVIS v1.0 dataset \cite{gupta2019lvis}. Following \cite{cui2019class}, we deﬁne the imbalance ratio (IR) of a dataset as the class size of the ﬁrst head classes divided by the size of the last tail classes in the training set. The details of these datasets are present in the Supplemental.

For classification, after training on the long-tailed training sets, we evaluate the models on the balanced validation/test sets and report top-1 recognition accuracy over all classes. To examine performance variations across classes with different examples numbers during training, we follow ~\cite{liu2019large} to report accuracy on three splits of these classes: Many-shot (more than 100 images), Medium-shot (20-100 images) and Few-shot (less than 20 images). For object detection and instance segmentation, we train on the LVIS training set and report mean Average Precision (mAP) on the validation set for all classes as well as three split groups including rare (appears in $\leq$ 10 images), common (appears in 10$-$100 images) and frequent (appears in $>$ 100 images)), respectively.

\begin{table}[]
\caption{Classification accuracy on Places-LT. $\dagger$ denotes results from \cite{liu2019large}.}
\begin{center}
\begin{tabular}{c c c c c}
\toprule
Methods & Few & Medium & Many & All \\ 
\midrule
$\text{Focal}^{\dagger}$\cite{lin2017focal} & 22.4 & 34.8 & 41.1 & 34.6   \\  
$\text{OLTR}^{\dagger}$\cite{liu2019large} & 25.3 & 37.0 & \textbf{44.7}\cite{kang2020decoupling} & 35.9 \\ \midrule
$\text{NCM}$\cite{kang2020decoupling} & 27.3 & 37.1 & 40.4 & 36.4 \\ 
$\text{cRT}$\cite{kang2020decoupling} & 24.9 & 37.6 & 42.0 & 36.7 \\ 
$\tau\text{-norm}$\cite{kang2020decoupling}  & 31.8 & \textbf{40.7} & 37.8 & 37.9 \\ 
$\text{LWS}$\cite{kang2020decoupling} & 28.6 & 39.1 & 40.6 & 37.6 \\ 
$\text{BALMS}$\cite{ren2020balanced} & 31.6& 39.8& 41.2& 38.7 \\ \midrule
Baseline   & 7.9 & 25.9 & 43.7 & 28.8 \\  
(Ours) LDA   & \textbf{32.1} & \textbf{40.7} & 41.0 & \textbf{39.1}  \\ 
\bottomrule
\end{tabular}
\end{center}
\label{tab:place}
\end{table}
  
\subsection{Implementation Detail}
\textbf{ImageNet-LT.} The model is trained for 90 epochs in total from scratch. Following~\cite{kang2020decoupling}, ResNeXt-50-32x4d~\cite{xie2017aggregated} is set as the backbone network. The SGD optimizer with momentum 0.9, batch size 256, cosine learning rate schedule decaying from 0.2 to 0 and image resolution 224$\times$224 is used. The $\gamma$ parameter in Equation \ref{eq:alpha} is set as 2.0. Unless otherwise specified, the warm-up strategy is used in first 5 epochs. 

\noindent\textbf{Places-LT.} Following~\cite{kang2020decoupling}, ResNet-152  pre-trained on ImageNet is set as the backbone network. During training, weights of the last block changes while other weights are frozen. $\gamma$ and $\beta$ are respectively set as 0.6 and 0.5.

\noindent\textbf{CIFAR100-LT.} Following~\cite{zhou2020bbn}, ResNet-32 is set as the backbone network.
The network is trained for 200 epochs and the SGD optimizer has a momentum 0.9, batch size 128 and base learning rate 0.1. The learning rate is decayed at 160-th and 180-th epoch by 0.01. $\gamma$ and $\beta$ are respectively set as 1.0 and 0.5. 

\noindent\textbf{LVIS.} We equip the Mask R-CNN with three classifiers (present in the Supplemental), as in Figure~\ref{fig-flowchart}. For training, we apply repeat factor sampling ~\cite{gupta2019lvis}, scale jitter (sampling image scale for the shorter side from 640, 672, 704, 736, 768, 800) and random flip at training time. For testing, images are resized to a shorter image edge of 800 pixels and no test-time augmentation is used. The SGD optimizer has a momentum 0.9, weight decay 0.0001, batch size 8 and base leaning rate 0.01. $\gamma$ and $\beta$ are respectively set as 1.5 and 0.1. The learning rate is dropped by a factor of 10 at both 8-th(16-th, 27-th) and 11-th(22-th, 33-th) epoch when total epochs is 12(24, 36).

\begin{table*}[ht]
\caption{Performance comparison on the LVIS val set with ResNet-50. $\dagger$ denotes data from \cite{ren2020coco+}.}
\begin{center}
\begin{tabular}{c|c|c| c c c |c}
\toprule
Methods & Epochs & $\text{AP}_{bbox}$  & $\text{AP}_{r}$ & $\text{AP}_{c}$ & $\text{AP}_{f}$  & $\text{AP}_{mask}$ \\ \midrule
$\text{Mask R-CNN}$\cite{he2017mask}& 12& 22.5  & 9.6& 21.0& 27.8 & 21.7\\
$\text{cRT}^{\dagger}$\cite{kang2020decoupling}& 25+10& 24.8 & 14.7 & 22.5 & 27.8 & 23.2 \\
$\text{LWS}^{\dagger}$\cite{kang2020decoupling}& 25+10& 24.8 & 14.9 & 22.5 & 27.8 & 23.3 \\
$\text{EQL}^{\dagger}$\cite{tan2020equalization}& 25+10& 24.2 & 16.5 & 24.2 & 27.5 & 24.2 \\
Balanced sigmoid\cite{ren2020coco+}& 25+10& \textbf{26.7} & 18.3 & 24.0 & 29.3 & 25.1 \\ \midrule
Baseline & 12& 22.5& 10.2& 20.8& 28.0& 21.8\\
(Ours) LDA & 12& 25.2& 18.2&23.7 & 27.9 & 24.4 \\
(Ours) LDA & 24& 26.6 & 18.1 & \textbf{25.4} & 29.4 & 25.7 \\
(Ours) LDA & 36& \textbf{26.7} & \textbf{18.4} & 25.3 & \textbf{29.8} & \textbf{25.9} \\
\bottomrule
\end{tabular}
\end{center}
\label{tab:lvis}
\end{table*}

\subsection{Main Result}\label{results}
\textbf{Classification.}
We conduct experiments on ImageNet-LT, Place-LT and CIFAR100-LT, and report the results in Tables \ref{tab:imagenet}, \ref{tab:place} and \ref{tab:cifar100}, respectively. 
Additional results about backbone networks are presented in the Supplemental. One can see that LDA consistently outperforms the state-of-the-art methods ~\cite{kang2020decoupling, zhou2020bbn, tang2020long} for all datasets and backbone networks. On ImageNet-LT, it achieves 53.4\% accuracy, 1.6\% higher than that of De-confound-TDE\cite{tang2020long} and 3.5\% higher than that of LWS~\cite{kang2020decoupling}. On Place-LT, LDA achieves 39.1\% accuracy which is slightly higher than the state-of-the-art (BALMS~\cite{ren2020balanced}). On CIFAR100-LT, with a pre-trained model, LDA respectively achieves 2.1\%, 3.4\% and 3.3\% higher accuracy than De-confound-TDE\cite{tang2020long} when IR is 10, 50 and 100. These are significant margins for the challenging task.

\noindent\textbf{Object Detection and Instance Segmentation.}
Table \ref{tab:lvis} shows experiment results on the LVIS dataset, and further exploration can be found in the Supplemental. In Table \ref{tab:lvis}, when trained 12 epochs, LDA achieves 25.2 AP for object detection and 24.4AP for instance segmentation, which is 2.7 and 2.6 higher than the baseline method. When trained 24 epochs, LDA achieves 25.7 AP, which is 0.6 higher than the Balanced sigmoid method, which is implemented upon~\cite{kang2020decoupling}. 

\begin{figure}[t]
\begin{center}
 \includegraphics[width=1.0\linewidth]{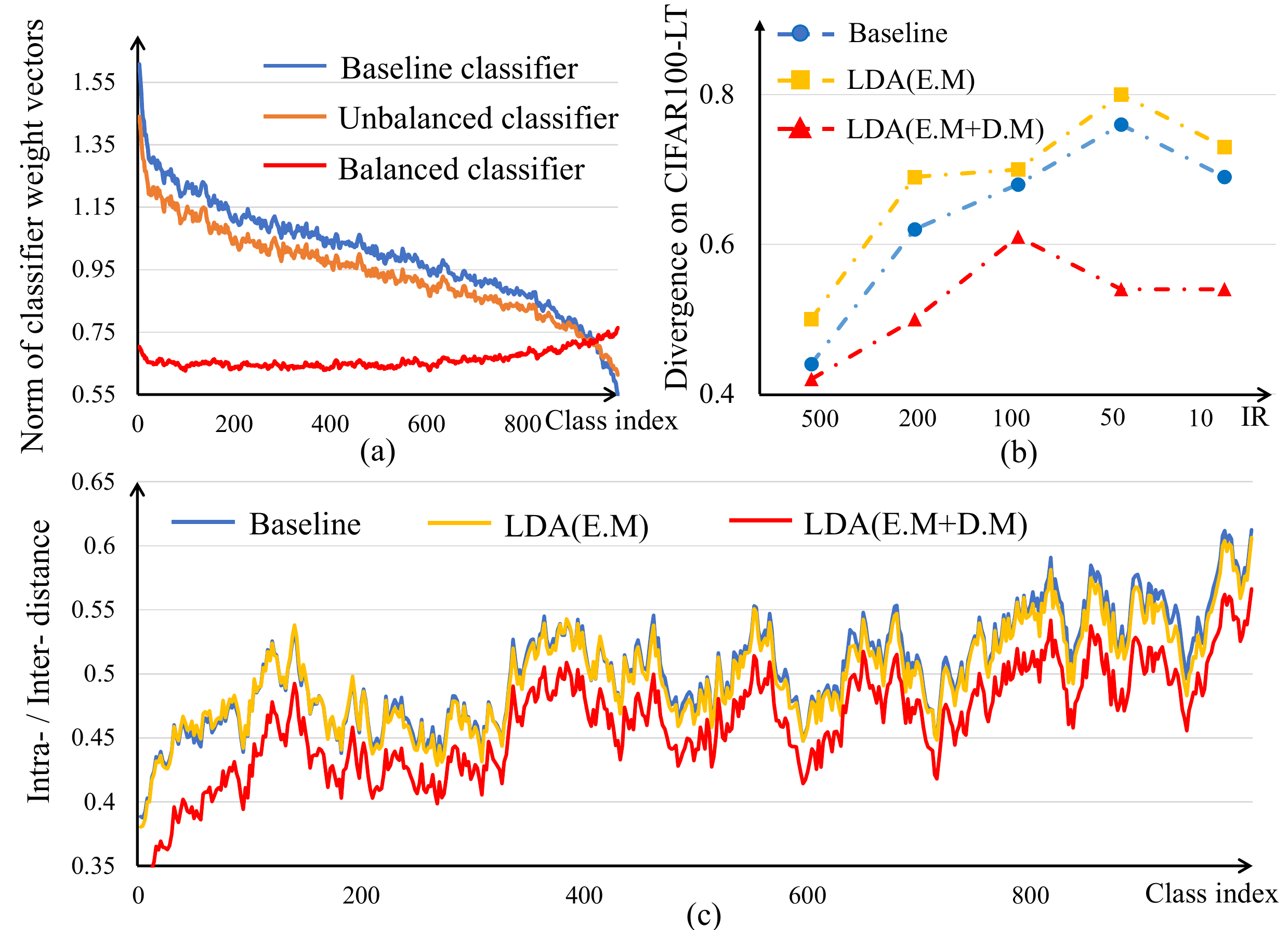}
\end{center}
 \caption{Visualization analysis. (a) Weight norm $||w_{c}||$ of classifiers. (b) Divergence between unbalanced and balanced domains. (c) Ratio between between the intra-class and inter-class distances ($d_{intra}^{c}/d_{inter}^{c}$). For short, the joint minimization of empirical risks and divergence minimization are denoted as E.M and D.M, respectively.}
\label{fig:norm-distance-divergence}
\end{figure}


\subsection{Visualization}

\textbf{Classifier Weight.}
The norm of the weight $||w_c||$ is correlated with the number of samples of class $c$, which is consist with the observation in~\cite{kang2020decoupling}. As shown in Figure \ref{fig:norm-distance-divergence}(a), the balanced classifier has uniform weight values for all categories, which demonstrate the bias of classifiers is largely solved by the proposed method. In comparison, weight values of the baseline classifier are significantly different for different categories.
As shown in Figure~\ref{fig:tsne_weight}, when trained with the plain cross entropy loss, the classifier weight vectors tend to mix together, which implies that the decision boundaries in the high-dimensional space are not clear. In contrast, weight vectors of the balanced classifier are more separated, which suggests the classifier is more discriminative.

\noindent\textbf{Domain Divergence.}
To verify the effectiveness of the proposed self-regularization strategy in Subsection~\ref{minimizingdivergence}, after inference phase, we calculate the contrastive domain divergence, which can reflects the intra-class and inter-class domain discrepancy for unsupervised domain adaptation. Please refer to ~\cite{kang2019contrastive} for more details. Figure~\ref{fig:norm-distance-divergence}(b) shows the divergence on CIFAR-100-LT with different imbalance ratios. 
The results show that, under different imbalance ratios, empirical risk minimization (E.M) slightly increases the divergence while divergence minimization (D.M) significantly decreases the divergence, which demonstrates the rationality and validity of the proposed self-regularization strategy.

\noindent\textbf{Feature Representation.}
To visualize the impact of LDA to feature representation, we measure the ratio of $d_{intra}$ to $d_{inter}$ on the test set, as shown in Figure~\ref{fig:norm-distance-divergence}(c). With LDA(E.M), the ratio almost keep the same as baseline. But with LDA (E.M+D.M), the ratio becomes significantly smaller, showing the higher representative power of features for the tail classes. 


\begin{figure}[!t]
\begin{center}
 \includegraphics[width=1.0\linewidth]{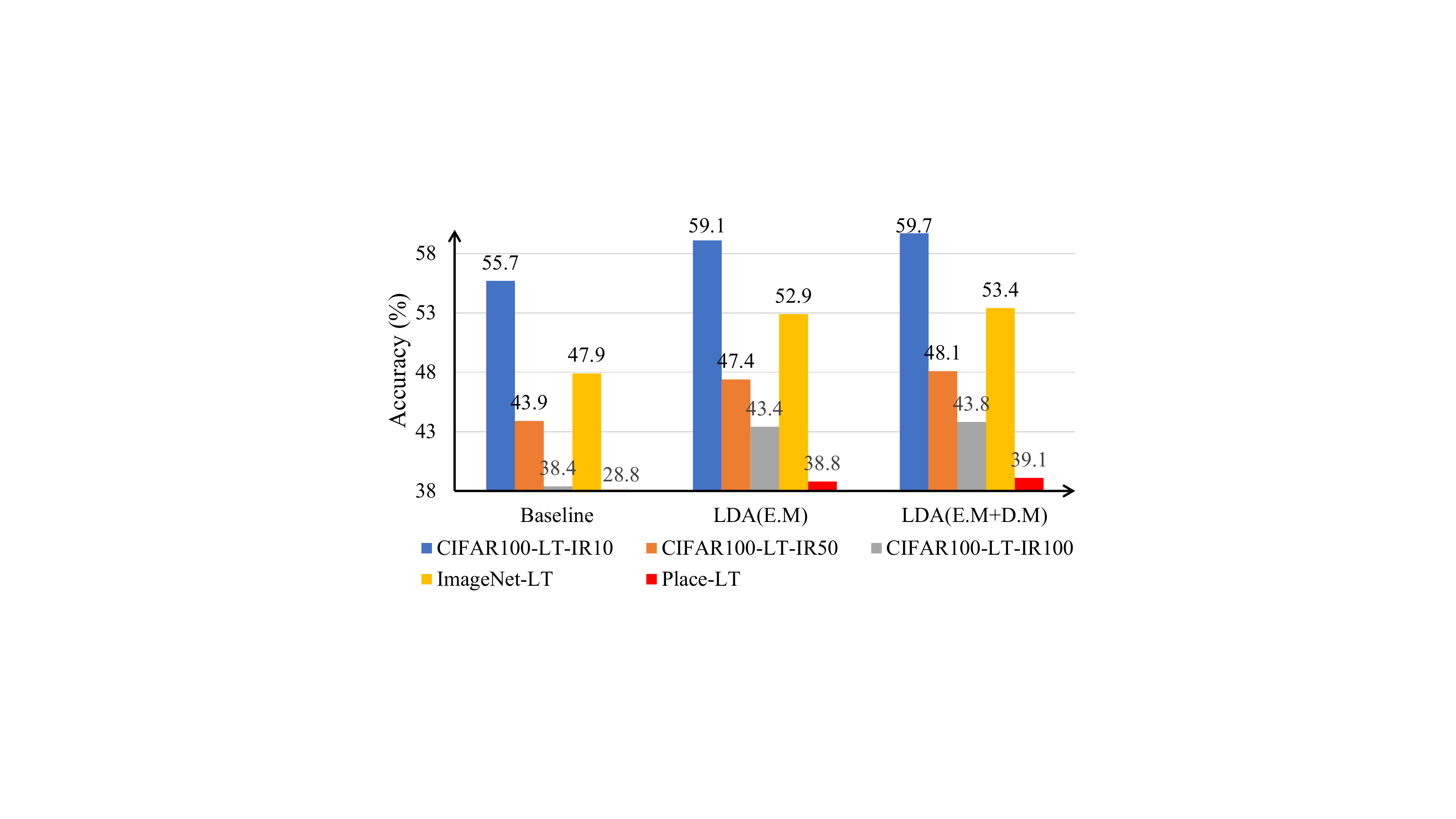}
\end{center}
 \caption{Analysis of E.M and D.M in LDA. E.M and D.M respectively denote empirical risk minimization and divergence minimization.}
\label{fig:ablaemdm}
\end{figure}

\subsection{Ablation Study}

\textbf{Empirical Risk and Domain Divergence.}
As shown in Figure \ref{fig:ablaemdm}, by using empirical risks minimization (E.M), LDA achieves 5\% improvement over the baseline method on ImageNet-LT, which demonstrates that combining risks from both the balanced and unbalanced domains is plausible. 
Moreover, with divergence minimization (D.M), LDA further improves the accuracy by 0.5\%, which not only confirms that the domain divergence is an important factor to be considered in long-tailed recognition but also validates the proposed divergence minimization strategy.


\noindent\textbf{Joint Optimization.}
As shown in Table~\ref{tab:ablajoint}, we compare optimization strategies. 
It can be seen that solely optimizing the emperical risk of the unbalanced ($\hat{\epsilon}_{u}$) or balanced domain ($\hat{\epsilon}_{b}$) achieves poor performance (41.3\%-47.9\%).
Optimizing the unbalanced ($\hat{\epsilon}_{u}$) and balanced domain ($\hat{\epsilon}_{b}$) step by step achieves much higher performance (52.1\%).
Such strategy has also been validated in ~\cite{kang2020decoupling} (CE$\rightarrow$RS) and in \cite{cao2019learning}(CE$\rightarrow$RW), which can be regarded as specific cases of the step-by-step optimization. Our joint optimization strategy ($\hat{\epsilon}_{u}\&\hat{\epsilon}_{u}$) achieves the best performance.

\begin{figure}[!t]
\begin{center}
 \includegraphics[width=1.0\linewidth]{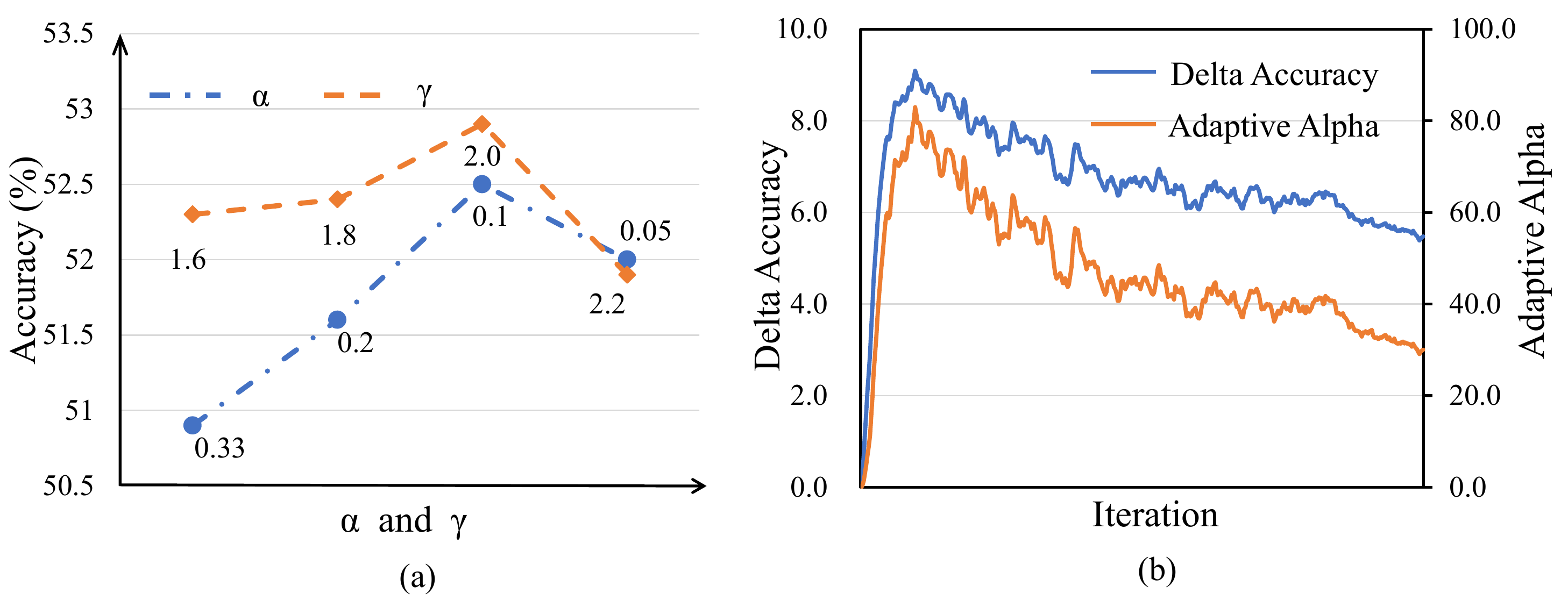}
\end{center}
 \caption{Ablation and visualization for self-adaptive strategy on ImageNet-LT. (a) Comparison of balance strategies. $\alpha$ is the balance factor generated by grid search to make a trade-off between empirical risks. $\gamma$ is a hyper-parameter defined in Equation \ref{eq:alpha}, (b) Change trend of adaptive alpha with the real-time performance difference $\Delta acc$.
 }
\label{fig:ablation-alpha}
\end{figure}

\noindent\textbf{Balance Strategy.}
To make a trad-off between the empirical risks of the unbalanced and balanced domains, we design the self-adaptive strategy based on the performance difference of unbalanced and balanced classifier. $\alpha$ in Equation~\ref{eq:alpha} is determined by the grid search method. In Figure \ref{fig:ablation-alpha}(a), we compare grid search with our self-adaptive adjustment strategy. It can be seen that our strategy outperforms the grid search method. When $\gamma$ is set as 2.0, LDA achieves the best performance.


\begin{table}[!t]
\caption{Comparison of optimization strategies. Top-1 accuracy of ResNeXt-50 on ImagetNet-LT test set. $\hat{\epsilon}_{u}$ and $\hat{\epsilon}_{b}$ denote optimizing unbalanced and balanced empirical risks, respectively. $\hat{\epsilon}_{u} \rightarrow \hat{\epsilon}_{b}$ means optimizing $\hat{\epsilon}_{u}$ and $\hat{\epsilon}_{b}$ step-by-step. $\hat{\epsilon}_{u}$\&$\hat{\epsilon}_{b}$ means joint optimization of $\hat{\epsilon}_{u}$ and $\hat{\epsilon}_{b}$. Ours-u and Ours-b respectively denote the unbalanced and balanced classifiers of the proposed LDA models.}
\begin{center}
\begin{tabular}{c c c c c c}
\toprule
Methods & $\hat{\epsilon}_{u}$ & $\hat{\epsilon}_{b}$ & $  \hat{\epsilon}_{u} \rightarrow \hat{\epsilon}_{b}$ & $ \hat{\epsilon}_{u}$\&$\hat{\epsilon}_{b}$ & accuracy \\ 
\midrule
CE   & $\checkmark$ &   &   &  & 47.9 \\  
RW   &   & $\checkmark$ &   &  & 41.3 \\ \midrule
CE$\rightarrow$RW   &   &  & $\checkmark$  &  & 51.9  \\ 
CE$\rightarrow$RS   &   &  & $\checkmark$  &  & 52.1 \\
\midrule
Ours-u   &   &   &  & $\checkmark$ & 48.7 \\ 
Ours-b  &   &   &   & $\checkmark$ & 52.9 \\  
\bottomrule
\end{tabular}
\end{center}
\label{tab:ablajoint}
\end{table}

\section{Conclusion}
In this paper, we re-visited the long-tailed problem from a domain adaptation perspective. 
In specific, after formulating Long-tailed recognition as Domain Adaption (LDA), we propose the slack the generalization error bound, which is defined upon the empirical risks of unbalanced and balanced domains the the divergence between them, which also has a great potential in explaining popular decoupling learning strategies.
By jointly minimizing unbalanced and balanced empirical risks and minimizing the domain divergence, we adapted the feature representation and classifiers trained on the long-tailed distributions to balanced distributions. Experiments on four benchmark datasets for image recognition, object detection, and instance segmentation demonstrated effectiveness of our proposed approach, in striking constant with the state-of-the-arts. 
In an explainable way, our research provides a fresh insight to the long-tail recognition problem in real-world scenarios.

\begin{acks}
This work was supported by National Natural Science Foundation of China (NSFC) under Grant 61836012, 61771447 and 62006216, the Strategic Priority Research Program of Chinese Academy of Sciences under Grant No. XDA27000000.
\end{acks}

\bibliographystyle{ACM-Reference-Format}
\balance
\bibliography{acmart.bib}

\appendix
\section{Dataset Details}

{\bf Long-tailed CIFAR-100:} 
This dataset is the long-tailed version of CIFAR-100 manually created in \cite{cui2019class}. The imbalance ratios used in experiments are 100, 50 and 10 following existing works~\cite{liu2019large, zhou2020bbn, tang2020long, cui2019class, ren2020balanced}.

{\bf ImageNet-LT:} To construct an unbalanced real-world scene dataset, a long-tailed version of ImageNet~\cite{russakovsky2015imagenet} termed ImageNet-LT was created~\cite{liu2019large}. The details about how to sample data from ImageNet can be found in~\cite{liu2019large}. The  ImageNet-LT dataset has 115.8K training images about 1,000 classes of images, with an imbalance ratio 1280/5. The validation/test set has 20/50 images per class, following a uniform distribution.

{\bf Places-LT:} A long-tailed version of Places \cite{zhou2017places} is introduced in \cite{liu2019large} with an imbalance ratio 4980/5, which is more challenging than the ImageNet-LT. The validation/test set has 20/100 images per class, following a uniform distribution.

{\bf LVIS:} LVIS v1.0 (LVIS for simplicity) is a long-tailed dataset for object detection and instance segmentation~\cite{gupta2019lvis}. The images are sampled from the MS COCO dataset~\cite{lin2014microsoft}. Compared with COCO, LVIS contains approximately 160k images from 1203 classes with 2M instance annotations. It also expanded the validation set from 5k images to 20k images. Compared with the datasets for image classification, LVIS has a significantly larger imbalanced ratio 50552/1.

\section{Object Detection and Instance Segmentation}\label{dedetails}

\begin{figure}[!t]
\begin{center}
 \includegraphics[width=0.8\linewidth]{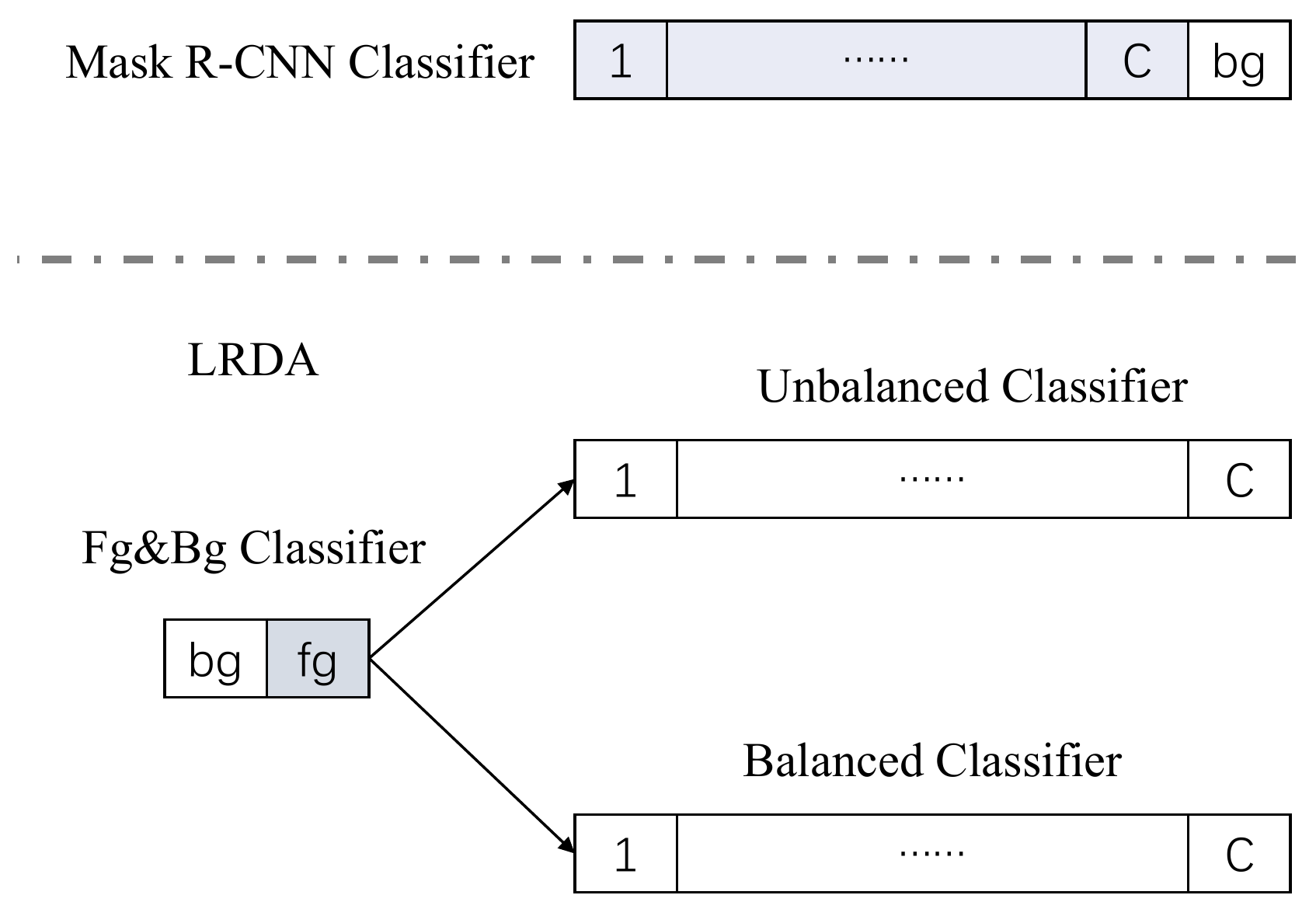}
\end{center}
 \caption{The classifier of our LDA in object detection and instance segmentation.}
\label{fig:mask_classifier}
\end{figure}

We choose Mask R-CNN~\cite{he2017mask} as the baseline for object detection and instance segmentation, following the settings in~\cite{liu2019large}. 
The head of Mask R-CNN has three branches, $i.e.,$ a classifier, a regression module, and a segmentation module. 
In detector implementation, the a regression module and segmentation module are kept as they are.
The classifier is updated to three classifiers: the first for Fg and Bg classification, the  second and the third are unbalanced and balanced classifiers, as shown in Figure \ref{fig:mask_classifier}. 
Following the long-tailed classification, we further use a fully connection layer (Figure 3 in the paper) for divergence minimization.

In Figure \ref{fig:mask_classifier}, Fg$\&$Bg is the foreground and background classifier, which is used to distinguish proposal is  foreground or not. 
All the prediction results of the proposal are used to calculate the loss for Fg$\&$Bg classifier.
The unbalanced and balanced classifiers are respectively used to minimize the empirical risks of unbalance and balance as defined in Subsection 3.2. 
Only the positive proposals are fed to the balanced and unbalanced classifiers.

During training, the loss function for Mask R-CNN is concluded by:
\begin{equation}
\begin{split}
    \mathcal{L} &= \mathcal{L}_{cls} + \mathcal{L}_{intra} + \mathcal{L}_{inter}+ \mathcal{L}_{box} + \mathcal{L}_{mask} \\
        &= \mathcal{L}_{Fg\&Bg} + \mathcal{L}_{LDA} + \mathcal{L}_{box} + \mathcal{L}_{mask}
\end{split}  
\label{eq:totallvisloss}
\end{equation}
Where $\mathcal{L}_{Fg\&Bg}$ denotes the cross entropy loss for the Fg$\&$Bg classifier, $\mathcal{L}_{LDA}$ denotes the loss defined in Equation 9 in the paper,
$\mathcal{L}_{box}$ and $\mathcal{L}_{mask}$ are kept same with~\cite{he2017mask}.

During inference, we multiple the score of Fg$\&$Bg with the score of the balanced classifier for object detection and instance segmentation.



\begin{table*}[t]
\caption{Results on ImageNet-LT with different backbone networks.}
\begin{center}
\begin{tabular}{c|c|ccc|c}
\toprule
Methods & Backbones & Few & Medium & Many & All  \\ \midrule

BALMS\cite{ren2020balanced} & \multirow{4}{*}{ResNeXt-10} & 25.3& 39.5& \textbf{50.3}& \textbf{41.8} \\
cRT\cite{kang2020decoupling}&                             & -   &   - &    -&  \textbf{41.8} \\
LWS\cite{kang2020decoupling}&                             & -   &   - &    -&  41.4 \\
LDA(Ours)                  &                             & \textbf{26.5}& \textbf{39.9}& 49.6& \textbf{41.8} \\ \midrule
cRT\cite{kang2020decoupling}& \multirow{4}{*}{ResNet-50}  & 26.1& 44.0& 58.8& 47.3 \\
LWS\cite{kang2020decoupling}&                             & 29.3& 45.2& 57.1& 47.7 \\
\cite{jamal2020rethinking}  &                             &    -&    -&    -& 48.0 \\
LDA(Ours)                  &                             & \textbf{31.2}& \textbf{48.4}& \textbf{61.1}& \textbf{50.9} \\ \midrule
cRT\cite{kang2020decoupling}& \multirow{3}{*}{ResNet-101} & 28.0& 46.5& 61.6& 49.8 \\
LWS\cite{kang2020decoupling}&                             & 31.2& 47.6& 60.1& 50.2 \\
LDA(Ours)                  &                             & \textbf{31.9}& \textbf{49.6}& \textbf{62.2}& \textbf{52.1} \\ \midrule
cRT\cite{kang2020decoupling}& \multirow{5}{*}{ResNeXt-101}& 27.0& 46.0& 61.7& 49.4 \\
LWS\cite{kang2020decoupling}&                             & 31.2& 47.2& 60.5& 50.1 \\
$\text{De-con}$\cite{tang2020long} &                      & 16.5& 44.3& \textbf{68.9}& 50.0 \\ 
$\text{De-con-TDE}$\cite{tang2020long} &                  & \textbf{33.0}& 50.0& 64.7& 53.3 \\ 
LDA(Ours)                  &                             & 32.5& \textbf{51.7}& 65.5& \textbf{54.4} \\ \bottomrule

\end{tabular}
\end{center}
\label{tab:varyimagenetbackbone}
\end{table*}

\begin{table*}[t]
\caption{Results on LVIS v1.0 val set with different backbone networks.}
\begin{center}
\begin{tabular}{c|c|c|cccc}
\toprule
Methods & Backbone & $\text{AP}_{bbox}$ & $\text{AP}_{r}$ & $\text{AP}_{c}$ & $\text{AP}_{f}$ & $\text{AP}_{mask}$ 
\\ \midrule
Mask R-CNN & \multirow{2}{*}{ResNet-50} & 22.5  & 9.6& 21.0& 27.8 & 21.7 \\
LDA     &                       & 25.2& 18.2& 23.7& 27.9 & 24.4 \\ \midrule
Mask R-CNN & \multirow{2}{*}{ResNet-101} & 24.6& 13.2& 22.7& 29.3 & 23.6 \\
LDA     &                       & 26.9 & 18.9 & 25.5 &  29.5& 25.9   \\ \midrule
Mask R-CNN & \multirow{2}{*}{ResNeXt-101-32x4d} & 26.7& 16.0& 24.8& 30.5 & 25.5 \\
LDA     &                       & 28.2& 20.1& 26.2&  30.7& 26.9  \\
\bottomrule
\end{tabular}
\end{center}
\label{tab:varylvisbackbone}
\end{table*}

\section{Compared Methods}\label{comparingmethods}
We compare our model with the following methods:

{\bf Cross-entropy loss.} We choose the vanilla cross-entropy loss as our baseline.

{\bf Label-distribution-aware margin loss.} Motivated by minimizing a margin-based generalization bound, It~\cite{cao2019learning} assigns different margins between classes according to classes frequency.  

{\bf Focal loss.} The focal loss\cite{lin2017focal} is a kind of hard example mining method, which aims to emphasize on learning hard examples by down-weight easy examples. It is proposed to tackle the foreground-background class imbalance in object detection. Therefore, it treats tailed class as hard sample in long-tailed classification.

{\bf Decoupling representation and classifier.} \cite{kang2020decoupling} decouples the learning pipeline into representation learning and classifier learning. Concretely, After learning representations with the random sampler, it adjusts the classifier by various methods(NCM, cRT, LWS or $\tau\text{-norm}$) with the class-balanced sampler.

{\bf Equalization loss.} \cite{tan2020equalization} analyzes the long-tailed recognition problem from the perspective of gradient, and tackles the problem by ignoring discouraging gradients for tail classes.

{\bf Bilateral-Branch Network.} \cite{zhou2020bbn} proposes a novel bilateral-branch architecture to simultaneously take care of the representation learning and classifier learning with a random sampler and a class-balanced sampler.

{\bf De-confound training and TDE.} \cite{tang2020long} establishes a causal inference framework to interpret what misleads the tail prediction biased towards head classes. It uses causal intervention in training, and counterfactual reasoning in inference.

\section{Backbone Networks}
To further validate the general applicability of the proposed LDA, we test it with different backbone networks for image classification, object detection and instance segmentation. The results are included in Tables \ref{tab:varyimagenetbackbone} and \ref{tab:varylvisbackbone}.

\end{document}